\documentclass{IEEEtran}
\usepackage{colortbl}

\usepackage{times}

\usepackage{soul}
\usepackage{url}
\usepackage[hidelinks]{hyperref}
\usepackage[utf8]{inputenc}
\usepackage[small]{caption}
\usepackage{graphicx}
\usepackage{amsmath}
\usepackage{booktabs}
\usepackage{color}
\usepackage{caption}
\usepackage{times}
\usepackage{helvet}
\usepackage{courier}
\usepackage{amsmath}
\usepackage{amssymb}
\usepackage{amsthm}
\usepackage{graphicx}
\usepackage{array}
\usepackage{booktabs}
\usepackage{amsopn}
\usepackage{amsmath,bm}
\usepackage{algorithm}
\usepackage{algorithmic}
\usepackage{enumerate}
\usepackage{color}
\usepackage{multirow}
\usepackage{bbding}
\usepackage{threeparttable}
\usepackage{threeparttable} 
\usepackage{dsfont} 
\usepackage{makecell} 

\title{Recent Advances in Vision Transformer: A Survey and Outlook of Recent Work}
\author{
Khawar Islam
\thanks{Khawar Islam is with FloppyDISK AI Research}
\thanks{Manuscript Submission June 14, 2022; revised August 00, 0000.}}

\begin{document}

\maketitle

\begin{abstract}
Vision Transformers (ViTs) are becoming more popular and dominating technique for various vision tasks, compare to Convolutional Neural Networks (CNNs). As a demanding technique in computer vision, ViTs have been successfully solved various vision problems while focusing on long-range relationships. In this paper, we begin by introducing the fundamental concepts and background of the self-attention mechanism. Next, we provide a comprehensive overview of recent top-performing ViT methods describing in terms of strength and weakness, computational cost as well as training and testing dataset. We thoroughly compare the performance of various ViT algorithms and most representative CNN methods on popular benchmark datasets. Finally, we explore some limitations with insightful observations and provide further research direction.
\end{abstract}

\section{Introduction}

Transformer \cite{vaswani2017attention} was introduced in the field of NLP on translating English-to-German language task. This transformer is solely based on attention mechanism, instead of convolution layers. It comprises of an encoder and decoder modules. Both modules are consists of attention layers with feed-forward network. Before, inserting into a transformer, every word in sentence embedded into vector $d_{model}=512$ dim. Recently, \cite{dosovitskiy2020image} applied existing architecture of transformer on computer vision task and achieved top performing results. Nowadays, ViTs have emerged and gave more promising results, can outperforms CNN. From an architectural point of view, the backbone of ViTs entirely made of self-attention mechanism and achieved remarkable success in visual tasks. There are two major concepts that significantly contributed towards the evolution of ViT models. (a)Self-attention mechanism where ViTs captured long-range of token dependencies in a global context, same as traditional recurrent neural networks. (b) The second important benefit is to train on large-scale unlabelled datasets and fine-tune on other tasks with small datasets. 

\begin{figure}[t]
	\centering
	\includegraphics[width=\linewidth]{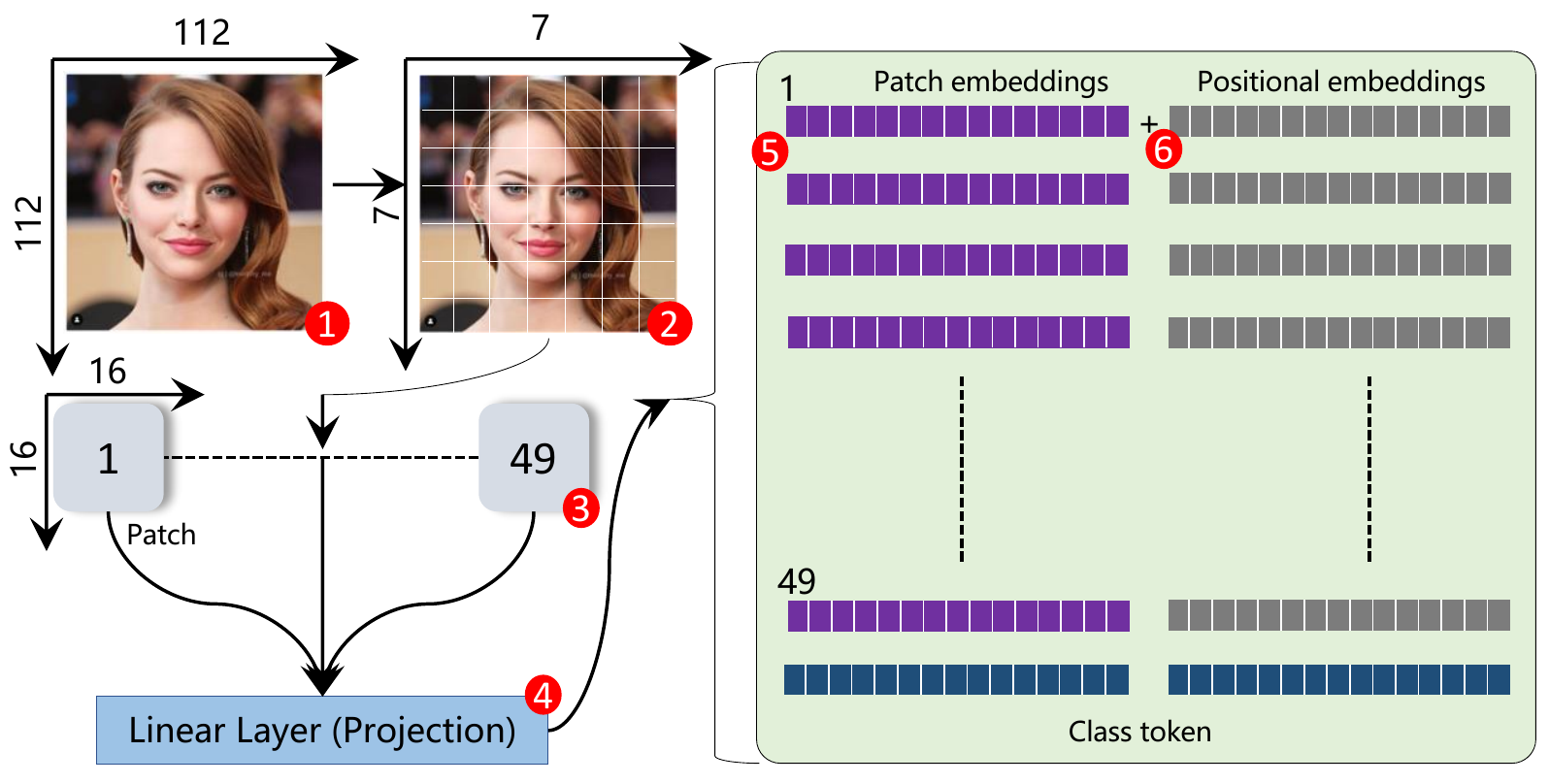}
	\caption{Additional steps of ViT.}
	\label{fig:ISVIT}
\end{figure}

\begin{figure*}[t]
	\centering
	\includegraphics[width=\linewidth]{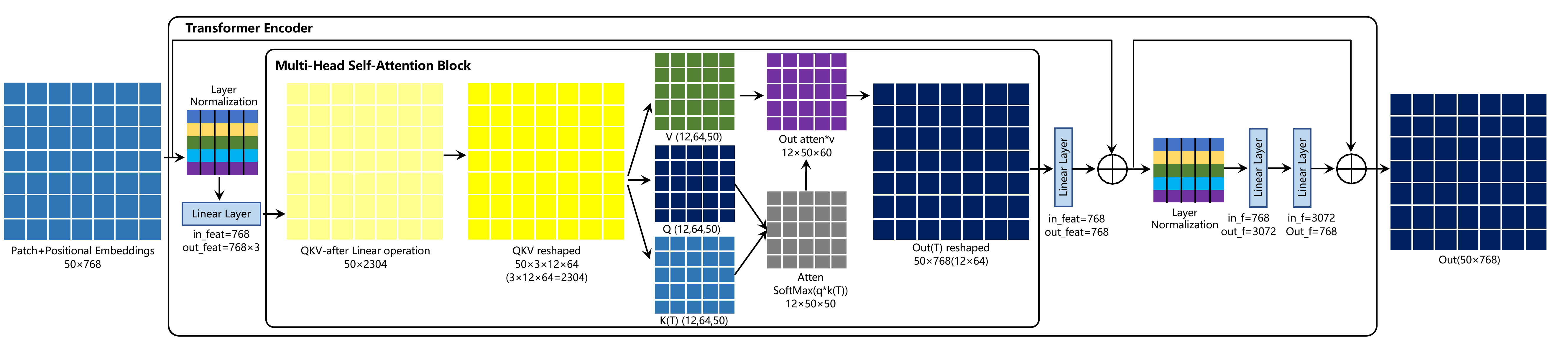}
	\caption{Overview of transformer encoder block in vision transformer along with multi-head self-attention module.}
	\label{fig:MHABLOCK}
	\vspace{-10pt}
\end{figure*}

\section{Overview of Vision Transformer} 
Graphical overview of additional ViT steps are presented in Figure. \ref{fig:ISVIT}. It requires several additional steps to utilize NLP transformer for computer vision tasks. The overall architecture of ViT can be presented into five main points:

\begin{itemize}
    \item Split image into non-overlap/overlap patches ($16\times16$, $32\times32$, etc.)
    \item Flatten patches and generate lower-dimensional linear embeddings from the flattened patches referred to (Patch Embedding).
    \item Add positional embedding and class token.
    \item Feed the sequence of patches into the transformer layer and obtain the output (label) through a class token.
    \item Pass class token values to MLP head to obtain final output prediction.
\end{itemize}
\paragraph{Step 1.} Consider an input image of size $112\times112$, we firstly generate $16\times16$ non-overlap/overlap patches. Thus, we can generate $49$ patches and straightforward feed into the linear projection layer. Remember that, the number of color channels in each patch is $3$. To obtain a long vector representation of each patch, the patches is fed into the projection layer and these representations have been presented in the Figure. \ref{fig:INITIALSTEPTANSFORMER} as 
\paragraph{Patch Embedding.} The total number of overlap/non-overlap patches are $49$, patch size with a number of channels is $16\times16\times3$. The size of the long vector of each patch is $768$. Overall, patch embedding matrix is $49$x$196$. Further, the class tokens have been added to the sequence of embedded patches and also added position Embedding. Without positional encoding, the transformer cannot retain the information and accuracy will be decreased to approx. 3\%. Now, the patch embeddings size becomes $50$ due to the additional class token. Finally, the patch embeddings with positional encoding and class token feed into the transformer layer and obtain the learned representations of the class token. Therefore, output from the transformer encoder layer is $1$x$768$ which is passed to the MLP block to obtain the final prediction.
\paragraph{Transformer Encoder Layer.} Especially in ViTs, the most important component is the transformer encoder that contains MHSA and MLP block. The encoder layer receives combined embeddings (patch embeddings, positional embeddings, and class tokens) of shape $50$ ($49$ patches and $1$ [cls] token)$\times768$($16\times16\times3$) as input. For all layers, the inputs and output of matrix shape $50$x$768$ from the previous layer. In ViT-Base architecture, there are $12$ heads (also known as layers). Before feeding input into the MHA block, the input is being normalized through the normalization layer in Fig. \ref{fig:MHABLOCK}. In MHA, the inputs are converted into $50\times2304$($768\times3$) shape using a Linear layer to obtain the Query, Key, and Value matrix. Then, reshaped these three QKV matrices into $50\times3\times768$ where each of the matrices of shape $50\times768$ presents the Query, Key, and Value matrices. Further, these Query, K, and Value matrices are reshaped again to $12\times50\times64$. Once, the Query, Key, and Value matrices are achieved, the attention operation inside the MHA module is performed using the below equation:
\begin{equation}
\mathrm{Attention}(\mathbf Q,\mathbf K,\mathbf V)=\mathrm{softmax}(\frac{\mathbf Q\cdot \mathbf K^\top}{\sqrt{d_k}})\cdot \mathbf V,
\end{equation}
Once we obtain the outputs from the MHSA block, these outputs are passed to skip connection as an input to obtain the final output. Before obtaining the final output, the output is fed into the normalization layer and passed to the MLP block. Originally, the MLP block was comprised of linear layers and a GELU activation function. Due to the significant progress in ViT, a locality mechanism \cite{li2021localvit} is introduced in MLP to capture local features. Furthermore, depth-wise convolution is embedded in the MLP block after the first FC layer to reduce parameters and achieve superior results. Finally, the outcome of the MLP block feeds into skip connection to obtain final outcome from an encoder layer.

\section{ViTs for Image Classification}
\begin{figure*}
    \centering
    \includegraphics[width=\textwidth,keepaspectratio]{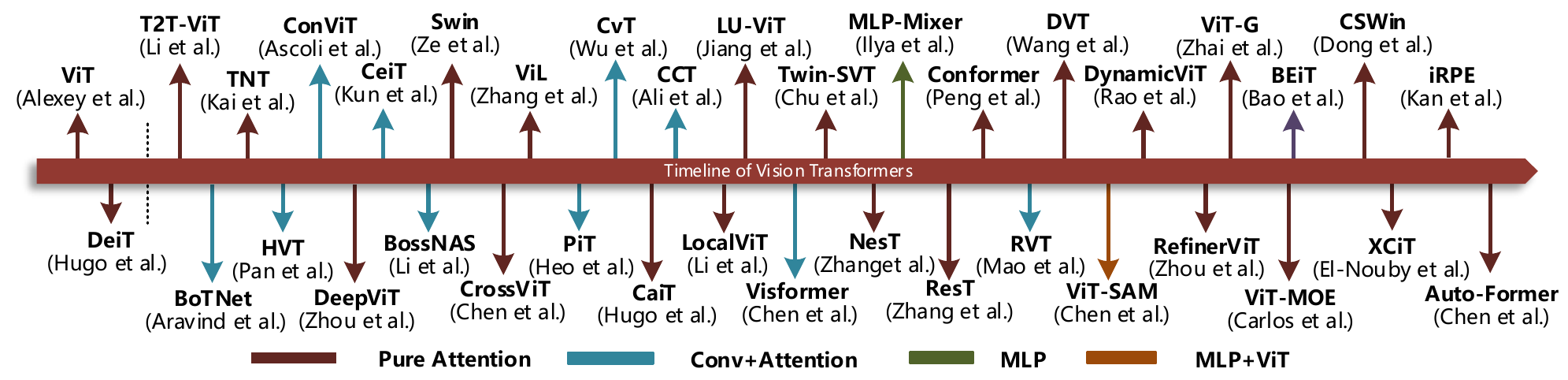}
    \caption{Timeline of existing vision transformer studies including (pure-attention, convolutional with attention, multi-layer perceptron, vision transformer with multi-layer perception, and self-supervised ViTs). As presented in the figure, the progress has started from the last quarter of $2020$. ViT \protect\cite{dosovitskiy2020image} and DeiT \protect\cite{touvron2020training} is available to the community for $2020$. In $2021$, a lot of vision transformers have been emerged enhancing training strategies, attention mechanism,s and positional encoding techniques. Major contributions to the Image-Net dataset for classification task is shown in this figure.}
    \vspace{-10pt}
     \label{fig:CLASSIFICATION_TIMELINE}
\end{figure*}

Recent development in ViTs have achieved valuable results in the field of classification tasks, several researchers have investigated further models for image classification tasks in different fields such as in medical images, generative images, etc (See in Table. \ref{tab:VIT_Methods}). Where transformers learn a useful representation from natural images. After a huge success of the top-performing transformers in NLP (e.g BERT, GPT, etc). \cite{dosovitskiy2020image} applied transformer for CV tasks and obtained SOTA results compare to conventional CNN and required less computation than CNN. To adjust the image as a sentence, the image is reshaped into 2D flattened patches. Then, learnable embeddings were added to embedded patches same as BERT class tokens. Finally, trainable positional embeddings are added into patch representation to preserve positional information. Transformers are totally dependent on the self-attention mechanism without the use of a convolutional layer. This self-attention technique helps the transformer to understand the relationship between image patches. In theory, an MLP performs better than a CNN model. But data has been a large obstacle with regard to the performance of MLP models. The timeline of ViT methods for image classification task is mentioned in Fig \ref{fig:CLASSIFICATION_TIMELINE}) \par

In addition to this, a distillation token has been added to the transformer, which interacts with image component tokens and classification vectors. In CNN architectures, we can easily increase performance by stacking more convolutional layers but transformers are different which are quickly saturated when architecture becomes deeper. The reason is that as the transformer enters the deep layer, the attention map becomes more and more similar. Based on this, \cite{zhou2021deepvit} introduced a re-attention module, which regenerated the attention map with a little computational cost in order to enhance the diversity between layers. Then, re-attention module trained on a 32-layer of ViT \cite{dosovitskiy2020image} transformer, the model achieved top-1 accuracy on Image-Net with the increment 1.6\% performance. \cite{touvron2020training} focused on the optimization part in transformers and optimized deeper transformers namely CaiT. It is similar to encoder and decoder where \cite{touvron2021going} explicitly split transformer layers involved self-attention between patches, from class-attention layers which were dedicated to extracting the content of the processed patches into a single vector so that it could be fed to a linear classifier. \par
Prior to existing transformers \cite{touvron2020training,dosovitskiy2020image,carion2020end}, \cite{yuan2021tokens} focused on training ViT from scratch and re-designed the tokens-to-token process which is helpful to simultaneously model images based on local-structure information and global correlation. Moreover, the above method significantly reduced the depth and hidden layer dimension. \cite{chen2021crossvit} introduced dual-branch ViT to extract multi-scale feature representations and developed a cross-attention-based token-fusion mechanism, which is linear in terms of memory and computation to combine features at different scales.
\begin{table}
    \centering
    \renewcommand\arraystretch{1.2}
     \setlength{\tabcolsep}{1.0mm}
    \caption{A major contribution of vision transformers in different domains of computer vision applications.}
    \scalebox{0.75}{
\begin{tabular}{cll} 
\hline
\rowcolor[rgb]{0.929,0.929,0.929} \begin{tabular}[c]{@{}>{\cellcolor[rgb]{0.929,0.929,0.929}}l@{}} \textbf{Reference}\end{tabular} & \textbf{Applications} & \textbf{Highlights} \\ 
\hline
\cite{yang2021transformers} & Object detection & \begin{tabular}[c]{@{}l@{}}Fast training convergence, Multi-head cross-Atten\end{tabular} \\
\rowcolor[rgb]{0.933,0.933,0.933} \cite{wu2021generative} & \begin{tabular}[c]{@{}>{\cellcolor[rgb]{0.933,0.933,0.933}}c@{}}Generative video\end{tabular} & \begin{tabular}[c]{@{}>{\cellcolor[rgb]{0.933,0.933,0.933}}l@{}}Tackling quadratic cost, Next-frame prediction\end{tabular} \\
\cite{he2021transreid} & \begin{tabular}[c]{@{}c@{}}Object re-identification\end{tabular} & \begin{tabular}[c]{@{}l@{}}Pure vision transformer, Person and vehicle re-ID\end{tabular} \\
\rowcolor[rgb]{0.929,0.929,0.929} \cite{wang2021uformer} & \begin{tabular}[c]{@{}>{\cellcolor[rgb]{0.929,0.929,0.929}}c@{}}Image restoration\end{tabular} & \begin{tabular}[c]{@{}>{\cellcolor[rgb]{0.929,0.929,0.929}}l@{}}Local-enhanced window, Skip-connection schemes\end{tabular} \\
\cite{li2021trear} & \begin{tabular}[c]{@{}c@{}}~Action recognition\end{tabular} & \begin{tabular}[c]{@{}l@{}}Inter-frame attention, Mutual-attention fusion\end{tabular} \\
\rowcolor[rgb]{0.929,0.929,0.929} \cite{pan20213d} & \begin{tabular}[c]{@{}>{\cellcolor[rgb]{0.929,0.929,0.929}}c@{}}3D object detection\end{tabular} & \begin{tabular}[c]{@{}>{\cellcolor[rgb]{0.929,0.929,0.929}}l@{}}Point clouds backbone, Local and global context\end{tabular} \\
\cite{gao2021utnet} & \begin{tabular}[c]{@{}c@{}}Medical image segmentation\end{tabular} & \begin{tabular}[c]{@{}l@{}}Convolution for features Long-range association\end{tabular} \\
\rowcolor[rgb]{0.929,0.929,0.929} \cite{du2020vtnet} & \begin{tabular}[c]{@{}>{\cellcolor[rgb]{0.929,0.929,0.929}}c@{}}Object goal navigation\end{tabular} & \begin{tabular}[c]{@{}>{\cellcolor[rgb]{0.929,0.929,0.929}}l@{}}Object instances in scene, Spatial locations, regions\end{tabular} \\
\cite{chen2021transformer} & Tracking & \begin{tabular}[c]{@{}l@{}}Template, search region, Ego-context block\end{tabular} \\
\rowcolor[rgb]{0.929,0.929,0.929} \cite{yuan2021volo} & \begin{tabular}[c]{@{}>{\cellcolor[rgb]{0.929,0.929,0.929}}c@{}}Visual recognition\end{tabular} & \begin{tabular}[c]{@{}>{\cellcolor[rgb]{0.929,0.929,0.929}}l@{}}Light-weight attention, Two-stage architecture\end{tabular} \\
\cite{yuan2021volo} & \begin{tabular}[c]{@{}c@{}}Lane shape prediction\end{tabular} & \begin{tabular}[c]{@{}l@{}}Non-local interactions, Directly regressed output\end{tabular} \\
\rowcolor[rgb]{0.929,0.929,0.929} \cite{zheng2021rethinking} & \begin{tabular}[c]{@{}>{\cellcolor[rgb]{0.929,0.929,0.929}}c@{}}Semantic segmentation\end{tabular} & \begin{tabular}[c]{@{}>{\cellcolor[rgb]{0.929,0.929,0.929}}l@{}}Sequence-to-sequence, Different decoder designs\end{tabular} \\
\cite{hudson2021generative} & \begin{tabular}[c]{@{}c@{}}Generative adversarial\end{tabular} & \begin{tabular}[c]{@{}l@{}}Bipartite structure, High-resolution synthesis\end{tabular} \\
\rowcolor[rgb]{0.929,0.929,0.929} \cite{yang2021transformers} & \begin{tabular}[c]{@{}>{\cellcolor[rgb]{0.929,0.929,0.929}}c@{}}Pixel-wise prediction\end{tabular} & \begin{tabular}[c]{@{}>{\cellcolor[rgb]{0.929,0.929,0.929}}l@{}}Gates attention approach, Multi-scale information\end{tabular} \\
\cite{coccomini2021combining} & \begin{tabular}[c]{@{}c@{}}Deepfake detection\end{tabular} & \begin{tabular}[c]{@{}l@{}}Mixed architecture, Local and global details\end{tabular} \\
\rowcolor[rgb]{0.929,0.929,0.929} \cite{zhang2021styleswin} & \begin{tabular}[c]{@{}>{\cellcolor[rgb]{0.929,0.929,0.929}}c@{}}HR image generation\end{tabular} & \begin{tabular}[c]{@{}>{\cellcolor[rgb]{0.929,0.929,0.929}}l@{}}Global positioning, Double attention\end{tabular} \\
\cite{park2021fast} & \begin{tabular}[c]{@{}c@{}}3D point cloud\end{tabular} & \begin{tabular}[c]{@{}l@{}}Light-weight local-Attenation, 136 times faster inference\end{tabular} \\
\rowcolor[rgb]{0.929,0.929,0.929} \cite{zou20216d} & \begin{tabular}[c]{@{}>{\cellcolor[rgb]{0.929,0.929,0.929}}c@{}}6D object pose estimation\end{tabular} & \begin{tabular}[c]{@{}>{\cellcolor[rgb]{0.929,0.929,0.929}}l@{}}Pixel-wise appearance, Point-wise geometric\end{tabular} \\
\cite{li2021danceformer} & \begin{tabular}[c]{@{}c@{}}3D dance generation\end{tabular} & \begin{tabular}[c]{@{}l@{}}Cascading kinematics, Phantom Dance dataset\end{tabular} \\
\rowcolor[rgb]{0.929,0.929,0.929} \cite{sun2021ssat} & \begin{tabular}[c]{@{}>{\cellcolor[rgb]{0.929,0.929,0.929}}c@{}}Makeup transfer and removal\end{tabular} & \begin{tabular}[c]{@{}>{\cellcolor[rgb]{0.929,0.929,0.929}}l@{}}Semantic correspondence, Semantic alignment\end{tabular} \\
\cite{li2021sctn} & \begin{tabular}[c]{@{}c@{}}Scene flow estimation\end{tabular} & \begin{tabular}[c]{@{}l@{}}Irregular point cloud, Rich contextual information\end{tabular} \\
\rowcolor[rgb]{0.929,0.929,0.929} \cite{fan2021svt} & \begin{tabular}[c]{@{}>{\cellcolor[rgb]{0.929,0.929,0.929}}c@{}}Place recognition\end{tabular} & \begin{tabular}[c]{@{}>{\cellcolor[rgb]{0.929,0.929,0.929}}l@{}}Super light-weight network, Long-range contextualdata\end{tabular} \\
\cite{bai2021towards} & \begin{tabular}[c]{@{}c@{}}Image Compression\end{tabular} & \begin{tabular}[c]{@{}l@{}}Lightweight image encoder Long-term information\end{tabular} \\
\rowcolor[rgb]{0.929,0.929,0.929} \cite{he2021transfg} & \begin{tabular}[c]{@{}>{\cellcolor[rgb]{0.929,0.929,0.929}}c@{}}Fine-grained recognition\end{tabular} & \begin{tabular}[c]{@{}>{\cellcolor[rgb]{0.929,0.929,0.929}}l@{}}Part selection module, Integrate raw Atten weights\end{tabular} \\
\cite{gong2021ssast} & \begin{tabular}[c]{@{}c@{}}Audio and speech\end{tabular} & \begin{tabular}[c]{@{}l@{}}Audio Spectrogram, Generative pre-training\end{tabular} \\
\rowcolor[rgb]{0.929,0.929,0.929} \cite{tian2021cctrans} & \begin{tabular}[c]{@{}>{\cellcolor[rgb]{0.929,0.929,0.929}}c@{}}Crowd Counting\end{tabular} & \begin{tabular}[c]{@{}>{\cellcolor[rgb]{0.929,0.929,0.929}}l@{}}Multi-scale receptive fields, Extract semantic features\end{tabular} \\
\cite{liang2022vrt} & \begin{tabular}[c]{@{}c@{}}Video Restoration\end{tabular} & \begin{tabular}[c]{@{}l@{}}Joint motion estimation, Parallel feature warping\end{tabular} \\
\rowcolor[rgb]{0.929,0.929,0.929} \cite{woo2022explore} & \begin{tabular}[c]{@{}>{\cellcolor[rgb]{0.929,0.929,0.929}}c@{}}Video Grounding\end{tabular} & \begin{tabular}[c]{@{}>{\cellcolor[rgb]{0.929,0.929,0.929}}l@{}}Explore-and-match model, Multiple sentences at once\end{tabular} \\
\cite{yan2022multiview} & \begin{tabular}[c]{@{}c@{}}Video Recognition\end{tabular} & \begin{tabular}[c]{@{}l@{}}Multi-resolution context, Lateral connections\end{tabular} \\
\hline
\end{tabular}
}
    \label{tab:arch4}
\end{table}


\section{Transformers for Segmentation}
Segmentation methods have been greatly increased due to the large-margin success of deep learning methods. It always remains an essential and important topic in the computer vision community. It includes three fundamental segmentation types such as (semantic, instance, and panoptic segmentation). Recently, various transformer methods have shown outstanding results in the domain of segmentation.
\subsection{Semantic Segmentation}
\cite{zheng2021rethinking} introduced a sequence-to-sequence approach and replaced the conv-encoder with a pure transformer. Three different decoders are designed to perform pixel-wise segmentation using progressive sampling, multi-level feature aggregation, and a naive up-sampling approach. \cite{strudel2021segmenter} presented a convolution-free end-to-end trainable encoder and decoder that captures contextual information. The encoding portion is built on standard ViT and depends on the encoding of patches. Further, a point-wise linear decoder is applied and decoded with a mask transformer. \cite{xie2021segformer} presented SegFormer, a simple yet powerful method with lightweight MLP decoders. An encoder is based on a hierarchical structure that gives multi-scale features and does not require any positional encoding scheme. SegFormer gets rid of complex decoders combining local attention and global attention to gender representation. \cite{wu2021fully} proposed a fully transformer network which relies on pyramid group transformer encoder to progress learned hierarchical features, while reducing the computational cost of standard ViT. Also, feature pyramid transformer fused spatial-level information and semantic level of the pyramid group transformer.

\subsection{Medical Image Segmentation}
Medical image segmentation is an important task for the development of healthcare frameworks, particularly for disease diagnosis (e.g biomedical, multi-organ, skin lesion, pannuke and etc). The majority of medical image segmentation problems, the U-NET\footnote{\textbf{U-NET} architecture is a de-facto standard for Medical Image Segmentation} architectures have become the major standard and achieved top-performing results. Different methods incorporate CNN approach to the self-attention approach to learning long-range dependencies among patches and the convolutional layer is used for extracting feature maps. \cite{chen2021transunet} proposed a hybrid TransUNet for learning high-resolution spatial features from CNN and long-range dependencies from Transformers. TransUNet obtained top-performing performance compared to several FCN-based methods included CNN and self-attention approaches. Transformers are data-hungry and require massive pre-training datasets to achieve comparable results to CNN. Especially for training on small-scale datasets, \cite{valanarasu2021medical} introduced a gated position-sensitive axial attention approach that worked on smaller datasets. Further, they proposed a local-global training technique for transformers. The global part worked on the actual resolution of an image, and a local part operated on image patches.
\par 
For 3D medical image segmentation, \cite{hatamizadeh2021unetr} introduced a volumetric transformer and proposed a unique architecture UNETR. They redefined 3D segmentation as 1D sequence-to-sequence task and utilized a attention mechanism in an encoder to obtain contextual knowledge from patches. Extracted features from encoder is combined with a decoder using residual learning (e.g., like ResNet) at multiple resolutions to predict the segmentation outputs. Similarly to multi-scale architecture, \cite{zhang2021pyramid} proposed a multi-scale attention technique to capture long-range relation by worked on multi-resolution images. To extract features, CNN-based pyramid architecture is employed. An adaptive-partitioning technique is applied to maintain informative relations and accessed different receptive fields. Recently, various transformer studies have been utilized self-attention mechanism similar to ViT \cite{dosovitskiy2020image}. As a backbone, it is appealing study to employ hierarchical structure in ViT. Swin Transformer \cite{liu2021swin} is a backbone network for vision tasks based on shifted windows technique. \cite{cao2021swin} proposed a Swin-Unet, an encoder utilized self-attention technique from local to global and decoder up-sampled the input global features to correspond pixel-level segmentation prediction. Different from previous work, \cite{li2021medical} presented Segtran for endless receptive fields to contextualize features for transformer. Segtran is based on Squeeze-and-Expansion an approach that design for solving images tasks. Segtran can easily see the fine details and local images.

\begin{figure}
	\centering
	\includegraphics[width=0.92\linewidth]{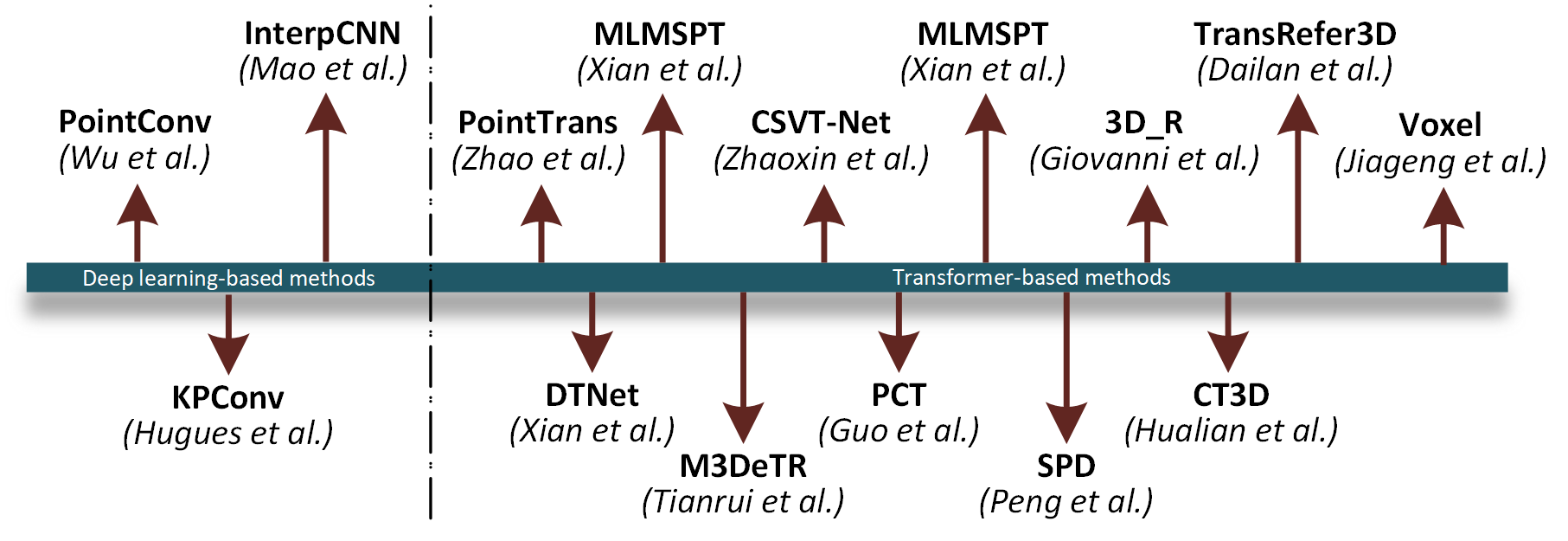}
	\caption{Point Transformer can serve as the backbone for various 3D point cloud tasks. \protect\cite{zhao2020point}}
	\label{fig:transformerlayer}
\end{figure}

\section{Transformers for 3D Point Clouds}

3D Point cloud aims to collect data in a space with rich geometric information, shape, and scale knowledge. Each data in point cloud data is represented as a 3D object or shape and contains Cartesian coordinates (X, Y, Z). The point collect data is collected from 3D sensors and devices or photogrammetry software. Recently, self-attention-based point clouds methods have been gained significant importance due to long-range dependencies, especially in this year. Different point cloud approaches have been proposed and achieved SOTA results compare to deep learning methods.
\begin{figure}
    \centering
    \includegraphics[width=\linewidth]{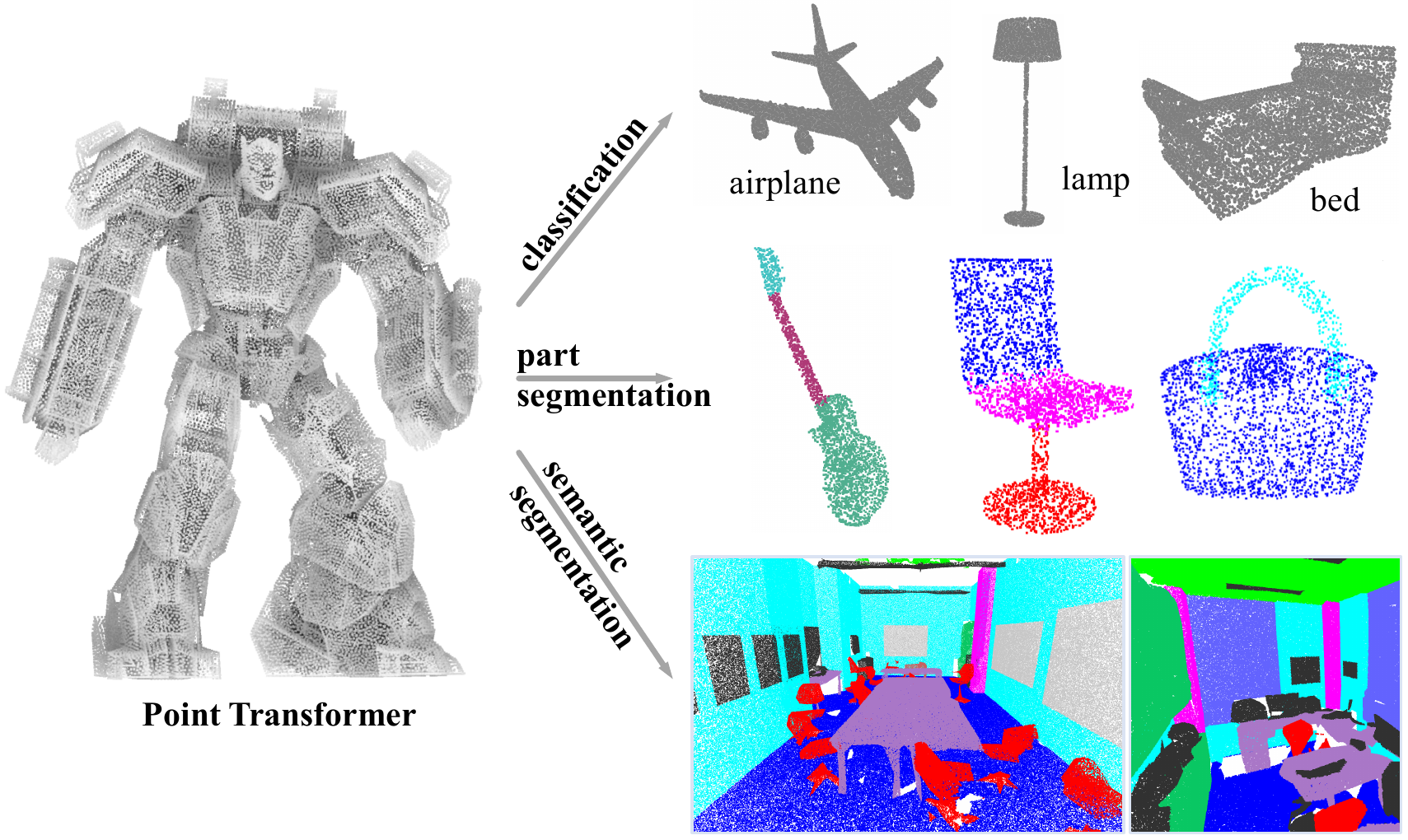}
    \caption{Top-performing method for point cloud registration (e.g., classification, part segmentation, and semantic segmentation.)}
    \label{fig:SOTA_POINT_CLOUD}
\end{figure}
Recently, \cite{zhao2020point} designed an impressive layer to serve as a backbone of dense prediction tasks and scene understanding. A proposed layer is invariant to permutation and cardinality and is therefore inherently suitable for the processing of point cloud data. By utilizing the proposed layer, they constructed a point transformer that is reliable to process point cloud data. Capturing long-range relationships is difficult in CNN networks, transformer solves the long dependencies problem through a self-attention mechanism. The transformer based SOTA method is highlighted in \ref{fig:SOTA_POINT_CLOUD}. Further, \cite{pan20213d} presented a feature learning-based backbone to capture global and local context. 
The main focus on the local region of the same instance then extends the attention process to other regions. Finally, attended to points from other instances globally, influencing both local and global relationships. \cite{han2021dual} developed a DPCT transformer network that can easily capture the long-range relationship and contextual knowledge using point-wise and channel-wise relationships. A unique multi-scale and multi-level point pyramid transformer \cite{han2021point} are designed to capture the cross-level and cross-scale context-aware features.

\section{Transformer for Person Re-Identification}

The goal of Re-ID is to retrieve a probe person from gallery images over various non-overlapping surveillance cameras \cite{islam2020person}. With the progressive development of DDNs and the rising requirement of high-tech video surveillance cameras, it has gained great interest in the CV and PR community. Re-ID has been broadly considered a topic in research as a particular person retrieval problem over non-overlapping CCTV cameras. Providing a probe image, the target of the Re-ID task is to identify whether this human has seemed in another place at a different time span captured by a different CCTV camera, or even the same camera at a different time instant. The probe person can be represented by an image, a video sequence, and even a text description. 
\par
Extracting discriminative features is a major and key problem in Re-ID. While CNNs based approaches have obtained remarkable success, they only process information via fixed-size grid mechanism at one time and lose important information details due to convolution and down-sampling operators (e.g. pooling and stride). Several transformer-based Re-ID methods have been designed to overcome these problems, \cite{he2021transreid} presented a strong baseline that utilized pure transformer and designed a jigsaw patches module, containing shift and patch shuffle operation, which promotes perturbation-invariant and robust discriminative features. Further, they introduced an SIE module that encoded side information via learnable embeddings. Xuehu et a. \cite{liu2021video} proposed a pure transformer-based Re-ID method to obtain rich features and extract in-depth video representation. Particularly, to jointly transform raw videos into spatial and temporal domains, a trigeminal feature extractor is proposed. Inspired by transformer architecture, three self-view transformers are designed to utilize token's dependencies and local information in ST domains. For extensive video representations, a cross-view method is developed to combine the multi-view features.
\par
\begin{figure}
    \centering
    \includegraphics[width=0.80\linewidth]{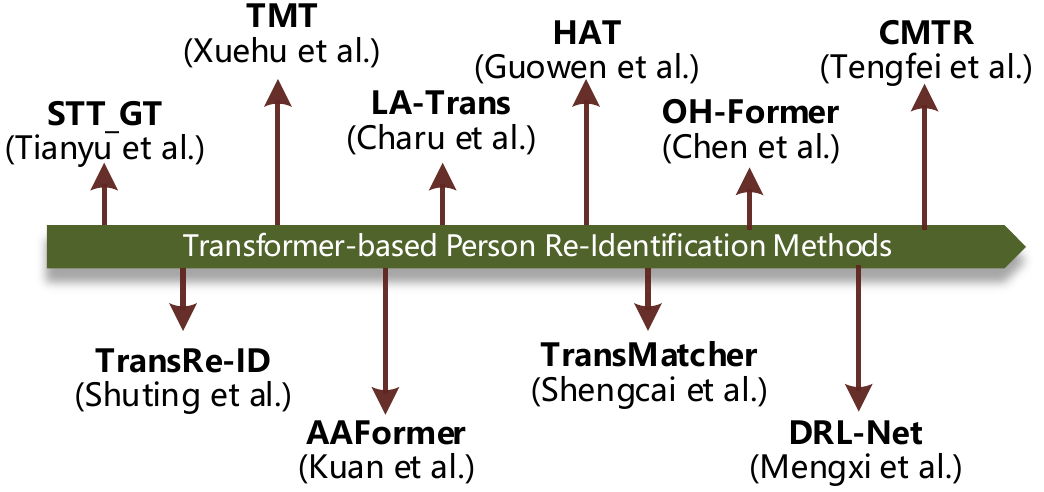}
    \caption{Timeline of self-attention-based mechanism methods for person re-identification.}
    \label{fig:vanilla_transformer}
\end{figure}
Different from Re-ID, video re-ID consists of continuous frames from surveillance cameras providing more rich spatio-temporal information that can be more effective and carefully note the movement of identity, view angle-changes, and appearance. \cite{zhang2021spatiotemporal} presented a simple transformer-based re-ID framework for the video re-ID task. Particularly, the constrained attention technique and global attention learning module are designed to utilize ST information from videos. To minimize the possibility of over-fitting, they introduced a synthesized data pre-training strategy to a better-initialized framework. In re-ID, extracting part-level features is a critical task to design a robust framework. \cite{zhu2021aaformer} introduced an alignment technique in transformer and presented a transformer to automatically find person and non-person parts at a single time at the patch level. Recently, in part-level works, \cite{sharma2021person} combined the output of global and local token of ViTs with a PCB-like technique to improve re-ID performance. They further incorporated a block-wise fine-tuning scheme to regularize the fine-tuned of a pre-trained backbone of ViTs. \cite{liao2021transformer, islam2021face, islam2022face111, islam2022face} applied a transformer for image matching and metric learning and proposed a simplified decoder for efficient image matching with an emphasis on similarity computation and mapping.  
\par
Combining CNNs and transformers is the new direction in the computer vision community and achieves more impressive results. Guowen e al. \cite{zhang2021hat} utilized CNNs and transformers to design a re-ID framework and proposed the DSA method to enhance the multi-scale features and combine hierarchical features from the CNN backbone. To integrate low-level information, transformer feature calibration is employed to obtain prominent knowledge and feed into each level of hierarchical features. Similarly, part-aware transformer \cite{li2021diverse} adopted CNN as a backbone network and proposed an occluded re-ID method. Encoder-decoder architecture is presented to utilize different portions of the human body. Part diversity and part discriminability are designed to obtain part-prototypes well with identity labels. \cite{jia2021learning} proposed disentangled representation network which handled occluded re-ID using global reasoning of local features without needing alignment of person image or further guidance. It's automatically measured image similarity of a person's body parts under the supervision of semantic knowledge of objects.

\subsection{Future Prospects}
Although recent years have observed remarkable progress in ViTs, there is still a lot of research that is hidden and unexplored. In this section, we highlight several new research directions for future work. \par
Most of the proposed ViT variants are highly computational expensive, require significant hardware, and are time-consuming. The first direction is to apply CNN pruning principle to prune ViTs according to their learnable scores. This technique will be helpful to reduce the complexity and memory footprint of ViT. In terms of the inference speed of ViT, pruning useless features, and preserving important features so that the final model can run more efficiently. Thus, deployment of the model becomes easier in real-time devices (e.g., smartphone, surveillance system) while maintaining the low resource cost. Therefore, it is a new research direction to obtain top-performing results while reducing the number of parameters and FLOPS.
\par
There are several unique qualities are present in pure ViT models like self-attention, global context, and generalization. While CNN is a widely used technique and popular for extracting pixel-level features (local features) and shared weights. Interestingly, the majority of studies have been investigated that a combination of CNN and ViT is more worthy instead of using straightforward pure ViTs. The main reason behind this, pure ViTs have lack shared weights, local context, and receptive fields. On other hand, CNN only works on pixel-level features, compressed all information through filters and does not consider the distanced pixel-relationship. Thus, it is an interesting topic to develop convolutional ViT or develop pure ViT without CNN, this question is still remains unsolved and more study required in this domain.
\par
ViT architectures have achieved superior performance compared to CNN in difficult tasks such as dense prediction and tiny object detection. These ViT models can easily able to learn an internal representation of the visual data which can surpass human capability. To some extent, an internal representation of the visual data remains secret in the architecture and keeps this technique as a black box. To explain how it works from an internal side, it is better to develop a new visualization layer that aims to understand how it works and be able to interpret its results. Thus, it is an interesting research area to develop an explainable vision transformer.
\par
Positional encoding is an important part of a transformer to capture a sequence of input tokens. However, positional encoding is not well investigated and studied, always remains a controversial topic which encoding scheme is best for it (e.g., conditional encoding or absolute encoding). Currently, the evidence available in the \cite{wu2021rethinking} relative positional encoding scheme is able to achieve top-performing results and focus on positional directivity. This question is for the future whether we need more positional encoding schemes or not.

\section{Conclusion}
ViTs are gaining more popularity and showing more impressive results in the domain of CV due to the advantage of the self-attention mechanism and creating a long-range relationship. In this chapter, we first describe the fundamental concepts of vision transformers and review the proposed ViT methods in recent years. Secondly, we highlight the influential work and draw a timeline of ViTs in different domains (e.g., classification, segmentation, point cloud, and person re-identification).
Lastly, we comparatively study different ViTs and CNN methods in terms of their accuracy on an image-Net dataset. Overall, ViT methods demonstrate excellent performance, and combining the self-attention mechanism with CNN shows outstanding results. Still, the training strategies and much-hidden work of ViTs have not been discovered yet and fully investigated.

\bibliographystyle{IEEEtran}
\bibliography{ijcai22.bib}

\end{document}